\documentclass{article} 
\usepackage{iclr2024_conference,times}

\usepackage{amsmath,amsfonts,bm}

\def\eqref#1{equation~\ref{#1}}

\def\1{\bm{1}}

\DeclareMathAlphabet{\mathsfit}{\encodingdefault}{\sfdefault}{m}{sl}
\SetMathAlphabet{\mathsfit}{bold}{\encodingdefault}{\sfdefault}{bx}{n}

\usepackage{hyperref}
\usepackage{url}

\usepackage{booktabs}
\usepackage{siunitx}
\usepackage{multirow}
\usepackage{graphicx}
\usepackage{amsmath}
\usepackage{wrapfig}

\title{Towards Unbiased Evaluation of Detecting Unanswerable Questions in EHRSQL}

\author{Yongjin Yang${}^{1}$\thanks{Equal contribution} \quad Sihyeon Kim${}^{1}$\footnotemark[1] \quad  SangMook Kim${}^{2}$\footnotemark[1] \quad  Gyubok Lee${}^{1}$ \vspace{2pt} \\ \textbf{Se-Young Yun}${}^{1}$ \quad \textbf{Edward Choi}${}^{1}$\thanks{Corresponding author}  \\ KAIST AI${}^{1}$ \qquad University of British Columbia${}^{2}$  \\
\texttt{\{dyyjkd, sihk, gyubok.lee, yunseyoung, edwardchoi\}@kaist.ac.kr}\\
\texttt{sangmook.kim@ece.ubc.ca}\\
}

\iclrfinalcopy 
\begin{document}

\maketitle

\begin{abstract}

Incorporating unanswerable questions into EHR QA systems is crucial for testing the trustworthiness of a system, as providing non-existent responses can mislead doctors in their diagnoses.
The EHRSQL dataset stands out as a promising benchmark because it is the only dataset that incorporates unanswerable questions in the EHR QA system alongside practical questions.
However, in this work, we identify a data bias in these unanswerable questions; they can often be discerned simply by filtering with specific N-gram patterns. 
Such biases jeopardize the authenticity and reliability of QA system evaluations.
To tackle this problem, we propose a simple debiasing method of adjusting the split between the validation and test sets to neutralize the undue influence of N-gram filtering.
By experimenting on the MIMIC-III dataset, we demonstrate both the existing data bias in EHRSQL and the effectiveness of our data split strategy in mitigating this bias.

\end{abstract}
\section{Introduction}
\label{sec:introduction}

Electronic Health Records\,(EHRs) are digital repositories that contain comprehensive medical information. 
EHRs help healthcare providers make informed decisions, reduce the likelihood of medical errors, and improve the overall healthcare quality.
Integrating a Question Answering\,(QA) system with EHRs introduces a valuable tool that can swiftly pinpoint critical information within the healthcare context.
Such a QA system can promptly respond to prevalent queries without the need for users to manually navigate through the entire EHRs.

While QA systems offer considerable convenience, they must be properly trained to provide accurate answers to the questions.
Additionally, to improve the reliability of QA system, it is vital for the system to decline unanswerable questions effectively.
The ability to distinguish unanswerable questions is crucial for trustworthiness as answering sensitive queries or providing inaccurate responses can mislead doctors and jeopardize patient care.

EHRSQL\citep{lee2022ehrsql} is a pioneering benchmark that uniquely integrates unanswerable questions into EHR QA validation and test sets with the aim of testing the reliability of the system. 
This incorporation significantly enhances the practicality of the data, setting it apart from existing EHR QA datasets \citep{wang2020text, raghavan2021emrkbqa}, as not all questions are answerable in real-world. 
Furthermore, EHRSQL enhances its practicality by introducing more realistic questions collected from doctors, nurses, and hospital administrative staff.
These distinctive features position EHRSQL as a valuable asset and a robust baseline for future research on EHR QA.

However, while exploring the unanswerable questions of EHRSQL dataset, we uncover a vulnerability arising from data bias: a notable quantity of unanswerable questions can be easily classified by simply filtering certain recurring phrases within them.
This data bias in unanswerable questions raises concerns about the reliability and stability of the dataset. The differentiation between answerable and unanswerable questions should not be straightforward\,\citep{rajpurkar2018know}, as this allows heuristic methods that do not understand the context to easily solve the problem.

In this paper, we delve into the data bias vulnerability of unanswerable questions within EHRSQL.
Our findings reveal that employing a simple heuristic approach, which uses N-gram-based filtering, can effectively detect numerous unanswerable questions.
When combined with the existing uncertainty-based method, this filtering approach improves the F1 score from $22.3$ to $93.2$.
To mitigate this vulnerability, we propose a new split of the validation and test sets, aiming to alleviate the inherent bias in the EHRSQL validation set. 
The primary motivation is to build a test set that includes questions with biased phrases, which are patterns unrecognized in the validation set.
Experiments on the MIMIC-III dataset show that this approach limits the heuristic use of N-gram patterns in yielding top results.
Consequently, the improved performance observed with new splits now more accurately reflects a true enhancement in the understanding of questions.


\vspace{-5pt}
\section{Data Bias in EHRSQL}
\label{sec:method}
\vspace{-0.2cm}




\begin{wraptable}{r}{6.5cm}
\vspace{-10pt}
\centering
{\small
\renewcommand{\arraystretch}{1.0}
\resizebox{0.45\columnwidth}{!}{

\begin{tabular}{clccc}
\toprule[1pt]
  Rank & N-gram Pattern  & \# ans & \# unans & ratio \\
\midrule
$\circ$ \ {\texttt{\,Unigram}} & & & & \\ 
(1) & department & 1 & 39 & 39 \\
(2) & you & 1 & 33 & 33 \\
(3) & appointment & 0 & 25 & 25 \\
(4) & can & 0 & 23 & 23 \\
(5) & phone & 0 & 21 & 21 \\
(6) & effects & 0 & 20 & 20 \\

\midrule
$\circ$ \ {\texttt{\,Bigram}} & & & & \\ 
(1) & other department & 0 & 20 & 20 \\
(2) & phone number & 0 & 19 & 19 \\
(3) & side effects & 0 & 18 & 18 \\
(4) & outpatient schedule & 0 & 18 & 18 \\
\midrule
$\circ$ \ {\texttt{\,Trigram}} & & & & \\ 
(1) & number of patient & 0 & 21 & 21 \\
(2) & the phone number & 0 & 16 & 16 \\
(3) & phone number of & 0 & 16 & 16 \\
\bottomrule[1pt]
\end{tabular}
}
}
\caption{Predominant N-grams in unanswerable questions of validation set.}
\vspace{-15pt}
\label{table:data_bias}
\end{wraptable} 

Data bias indicates patterns in datasets, potentially affecting model reliability. This section examines data bias in EHRSQL's unanswerable questions.



To identify data bias, we examine the unigram, bigram, and trigram features of unanswerable questions. First, we extract all the N-grams from both answerable and unanswerable questions in the training and validation sets. 
Next, the ratio of unanswerable-to-answerable questions is calculated for those N-grams. We then sort the N-grams based on this ratio to pinpoint patterns that predominantly appear in unanswerable questions.

As shown in Table \ref{table:data_bias}, certain patterns predominantly appear in unanswerable questions. 
For unigrams, the top six words that exhibit the highest ratio appear almost 150 times out of the 362 unanswerable questions. These limited patterns in the distinction of unanswerable cases enable simple heuristic methods to filter out those N-grams, thereby boosting performance.

Delving into the details of these N-grams, we can qualitatively explain why these words are absent from answerable questions. 
For example, based on the database schema of MIMIC-III, details about doctor appointments, departmental information, and phone numbers are not intended to be saved. 
Hence, it is evident that these phrases are often used to formulate unanswerable questions, making it easier to identify prevalent patterns in such questions.


\section{Mitigating Data Bias}

We have demonstrated the intrinsic data bias that exists within the EHRSQL. 
In this section, we offer a straightforward solution to address this bias by using a new split for the validation and test set. 
It is important to note that we make a new test set from the original validation set, as the original test set of EHRSQL is not accessible.

\begin{wrapfigure}{r}{5.5cm}
\vspace{-20pt}
    \centering
    \includegraphics[width=0.4\columnwidth]{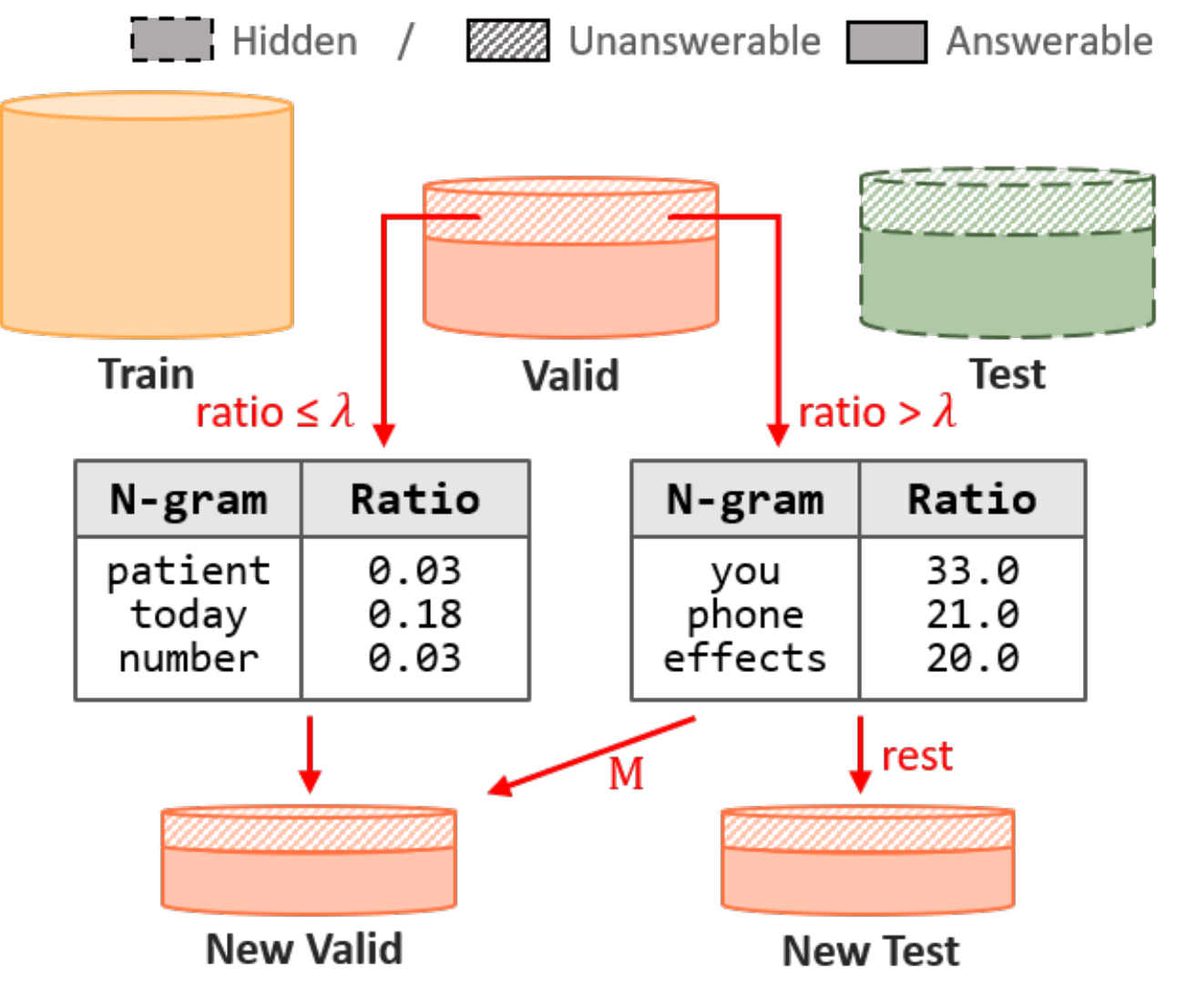}
    \vspace{-25pt}
    \caption{The overview of our proposed data split.}
    \label{fig:data}
    \vspace{-20pt}
\end{wrapfigure} 

Due to specific N-gram patterns in the unanswerable questions, filtering out questions with certain N-grams enables straightforward differentiation between answerable and unanswerable questions.
Relying on N-gram patterns can be detrimental to the task because they do not take into account the context of the questions. 
To address this issue effectively, one possible approach is to confine these N-gram patterns in the test set by removing them from the validation set.
Therefore, by creating a new test set primarily composed of questions exhibiting these patterns, we can significantly mitigate this bias.

Figure~\ref{fig:data} illustrates our approach to generate new data split.
We first estimate the predominant word ratio in the unanswerable questions of the validation data as explained in the Section~\ref{sec:method}. 
Next, we pinpoint questions that incorporate words with a high ratio that surpass the predefined threshold ratio denoted as $\lambda$, for both unigrams and bigrams. These identified questions are then allocated to the newly created test dataset. Moreover, to prevent a task from becoming cross generalization, a small fixed $M$ number of those questions for each N-gram with high ratio are transferred to a new validation set. 

The remaining unanswerable questions with a ratio below $\lambda$ are assigned to the new validation set.
Following this, we randomly split the answerable questions to the new validation set and test set, ensuring the elimination of the distinctive pattern associated with unanswerable questions.

\section{Experiments}
\label{sec:experiments}
\subsection{Experimental Setup}

\paragraph{Dataset} We use the MIMIC-III\,\citep{johnson2016mimic} dataset from EHRSQL for training and evaluation. 
The dataset contains a total of 9,318 training samples, all of which are answerable questions. The validation set has 1,122 samples, which includes 760 answerable questions and 362 unanswerable questions. 
There is also a hidden test set evaluated via the leaderboard website.

\paragraph{Implementation Details} We utilize the T5-base\,\citep{raffel2020t5} model. During the training phase, we set a learning rate to $0.0001$. For generation, we use beam decoding with a beam size of 5. 
For the construction of a new split, we set $\lambda$ to 20 for unigram, 16 for bigram, and $M$ is set at 5 to balance the number of data in each set.

\paragraph{Baselines} For comparison with baseline methods, we adopt two techniques known for measuring a model's uncertainty, which can subsequently be used for filtering: maximum entropy \,\citep{malinin2020uncertainty, xiao-wang-2021-hallucination} and beam score\,\citep{lopez2008beam}. Entropy serves as a measure of the uncertainty inherent in a language model, while the beam score is a metric used to select the most promising sequence. Detailed information with formula is presented in Appendix \ref{appendix:uncertainty_estimation}. If the entropy exceeds a predefined threshold or the beam score falls below one, we classify the question as unanswerable. 

\paragraph{N-gram Filtering}  
We combine baseline methods with the N-gram filtering, and test them on both original and new split datasets. This approach filters out the questions with specific patterns that are frequently found in unanswerable validation questions, thereby exploiting the data bias of EHRSQL.

As discussed in Section~\ref{sec:method}, we systematically identify all unigrams, bigrams, and trigrams within the validation set. We then calculate the ratio of occurrences between unanswerable and answerable questions for these N-grams and sort them based on this ratio. Our thresholds for filtering ratios are set at 8 for unigrams, 10 for bigrams, and 4 for trigrams in the original dataset. Specifically, this means that a question is filtered out if it contains specific unigrams occurring at least 8 times more frequently in unanswerable questions, bigrams occurring at least 10 times more, or trigrams occurring at least 4 times more. Additionally, baseline methods such as utilizing entropy are used to filter out unanswerable questions not caught by N-gram filtering. For the new split, thresholds are uniformly set at 6 for each N-gram category. These thresholds have been heuristically determined to maximize the target F1 score within constraints.
During the test phase, we utilize the same N-grams identified in the validation phase for filtering.

\paragraph{Evaluation} Following the evaluation approach of \citet{lee2022ehrsql}, our assessment is twofold. The metric $F1_{ans}$ evaluates the model's capability to determine whether a given question is answerable or not. We omit the $F1_{ans}$ metric for the EHRSQL test set since the leaderboard for the hidden test set does not provide it. Another crucial evaluation metric is $F1_{exe}$, which assesses the accuracy of the model's generated query by executing the produced SQL against the MIMIC-III database.
$F1_{exe}$ is calculated as a combination of precision\,($P_{exe}$) and recall\,($R_{exe}$).
$P_{exe}$ is the proportion of correctly answered questions among all questions predicted as answerable, while $R_{exe}$ represents the proportion of correctly answered questions among all questions categorized as answerable.
Furthermore, in the context of medical treatment, $P_{exe}$ is particularly critical because, if the system answers unanswerable questions, it provides incorrect responses to doctors. Doctors may then rely on this incorrect information to make decisions, which can be highly dangerous. In contrast, a low recall, which occurs if the model does not respond to unanswerable questions, is merely inconvenient. While this means doctors must seek the answers themselves, it does not critically impact patient health. Given these considerations, we place greater emphasis on $P_{exe}$ compared to $R_{exe}$ and mandate that the precision, $P_{exe}$, score exceeds 99.0 for the validation set. Regarding the threshold of 99.0, we follow the official EHRSQL leaderboard, which also emphasizes the importance of a high $P_{exe}$ value. Thresholds for entropy and beam score are established based on this criterion; when $P_{exe}$ exceeds 99.0, we select thresholds that yield the best balance of recall and $F1_{exe}$.
Appendix \ref{appendix:implementation_details} provides comprehensive implementation details.

\subsection{Results}

\vspace{-13pt}
\begin{table}[!ht]
\begin{minipage}[t]{.47\linewidth}
    \small
    \caption{Results on EHRSQL.}
\resizebox{\columnwidth}{!}{
    \begin{tabular}{lcccc|ccc} 
        \toprule
        \multirow{2}{*}{\textbf{Method}} & \multicolumn{4}{c|}{Valid} & \multicolumn{3}{c}{Test} \\
        \cmidrule{2-8} 
         & \textbf{$F1_{ans}$} & \textbf{$P_{exe}$} & \textbf{$R_{exe}$} & \textbf{$F1_{exe}$}  & \textbf{$P_{exe}$} & \textbf{$R_{exe}$} & \textbf{$F1_{exe}$} \\
         \midrule
         $Entropy$ & 7.1 & 100.0 & 3.7 & 7.1 &   100.0 & 3.3 & 6.5\\
         $Beam$ & 27.4 & 99.2 & 15.8 & 27.2 & 100.0 & 12.5 & 22.3 \\
         $Ngram$ & 7.8 & 100.0 & 4.1 & 7.8 &  92.3  & 3.0 & 5.8 \\
         $Entropy + Ngram$ & 70.2 & 99.0 & 53.6 & 69.5  & 94.5  &50.7  &66.0  \\
        $Beam + Ngram$ & 96.0 & 99.3 & 91.7 & {95.4} &   95.8 & 90.7 & {93.2} \\
         \bottomrule
    \end{tabular}
}
    \label{tab:main}
\end{minipage}\hfill%
\begin{minipage}[t]{.52\linewidth}
    \small
    \centering
    \caption{Results on the newly split EHRSQL.}
    \resizebox{\columnwidth}{!}{
    \begin{tabular}{lcccc|cccc} 
        \toprule
        \multirow{2}{*}{\textbf{Method}} & \multicolumn{4}{c|}{Valid} & \multicolumn{4}{c}{Test} \\
        \cmidrule{2-9} 
         & \textbf{${F1_{ans}}$} & \textbf{$P_{exe}$} & \textbf{$R_{exe}$} & \textbf{$F1_{exe}$} & \textbf{$F1_{ans}$} & \textbf{$P_{exe}$} & \textbf{$R_{exe}$} & \textbf{$F1_{exe}$} \\
         \midrule   $Entropy$ & 57.6 & 99.4 & 40.5 & 57.6 & 59.8 & 98.2 & 42.6 & 59.5 \\
         ${Beam}$ & 59.4 & 99.4 & 42.4 & 59.4 & 60.3 & 98.2 & 43.2 & 60.0 \\
         ${Ngram}$ & 2.1 & 100.0 & 1.1 & 2.1 &  1.6 & 100.0 & 0.8 & 1.6 \\
         $Entropy + Ngram$ & 57.3  & 99.4 & 40.3 & 57.3 & 59.6 & 98.2 & 42.4 & 59.2 \\
        $Beam + Ngram$ & 62.5 & 99.4 & 45.5 & 62.5 & 62.5 & 97.7 & 45.5 & 62.1 \\
         \bottomrule
    \end{tabular}
}
    \label{tab:split}
\end{minipage}
\end{table}
\vspace{-10pt}
\subsubsection{Results on EHRSQL}

Table \ref{tab:main} presents the results of each method on EHRSQL. Two primary observations emerge from the data. Firstly, the close alignment between $F1_{ans}$ and $F1_{exe}$ scores for validation set suggests that the model is predominantly accurate in generating the correct SQL query. This implies that the more formidable challenge is discerning between answerable and unanswerable questions. 
Second, the N-gram method's superiority over baselines is clear.
For instance, while the entropy-only method achieves an F1 score of $6.5$ on the test set, its combination with the N-gram technique elevates this score to $66.0$. A similar performance boost is evident with the beam score, where its integration with the N-gram approach lifts the F1 score by more than $70.0$. Although relying solely on the N-gram method falls short of impressive performance due to its ambition of achieving a $99.0$ precision score, its combination with other methods amplifies the overall performance. This suggests that the task has become overly simplistic and easily solvable with heuristic methods. Consequently, the effectiveness of combining any method to filter out unanswerable questions with N-gram filtering undermines the reliability of the benchmark. This is because such combinations work well regardless of the method's inherent efficacy, preventing an accurate assessment of the true capabilities of the model and the methods.

\subsubsection{Results with New Split}
Table \ref{tab:split} shows the results of each method on EHRSQL with the new data split. 
From the initial validation set, 558 samples are allocated to the new validation set and the remaining 564 samples form the new test set.
With this new split, it is clear that combining N-gram filtering with baseline methods does not always boost performance, and in some cases, it even reduces performance.
While removing certain patterns in the validation set increases the performance of baseline methods, they still struggle with an F1 score near $60.0$ demonstrating the task is still challenging. 
Moreover, relying solely on N-gram filtering yields poorer results with this new split, further emphasizing the reduction in data bias. 
As a result, our new split effectively counteracts data bias, providing a more accurate measure of a model's ability to refrain from responding to uncertain questions.
Appendix \ref{appendix:more_results} provides various ablation studies based on the new split.

\section{Conclusion}
In this study, we present evidence of data bias within EHRSQL through N-gram analysis. Our investigation shows that unanswerable questions in the validation dataset display a discernible pattern. By filtering the unanswerable questions with those N-grams and uncertainty measures, the task becomes easily solvable, casting doubt on its reliability. To address this, we introduce a partitioning strategy for both the validation and test datasets, relocating specific patterns to the test set to ensure they are not identified through validation set. 
Our research findings and experimental outcomes hold the promise of offering valuable insights into dataset design, aiming to bolster the reliability of EHR QA systems.
%
\section*{Limitations}


Although our method highlights the critical problem of data bias in the EHRSQL dataset and provides a simple solution to address it, our approach is specific to the EHRSQL data. This specificity is due to the fact that EHRSQL is the only dataset currently available that incorporates unanswerable questions, thus enabling the assessment of the reliability of trained EHR QA models. We look forward to investigating other EHR QA benchmarks that contain unanswerable questions and applying our solution in future work.

Additionally, although our solution effectively addresses the data bias in the EHRSQL dataset in some aspects, our proposed solution might not completely eliminate the inherent bias in the dataset, as some patterns still remain in unanswerable questions. It is essential to reiterate that our research primarily focuses on the vulnerabilities in EHRSQL data and proposes a straightforward strategy that does not require access to the hidden test set or the creation of a new dataset, which would involve significant cost and effort.

In the case of unanswerable questions, these often stem from categories such as side effects, upcoming examination schedules, attending physicians, patient personal information, patient consent status, and diagnoses received from other medical departments. To fully remove inherent bias, it's necessary to address vulnerabilities that arise during the categorization process. Annotators, especially doctors, tend to formulate unanswerable questions based on their past experiences, which can limit variation and introduce specific keywords into the datasets. To mitigate this issue, one approach could be to align the N-gram distribution between answerable and unanswerable questions. For example, a potential solution might involve crafting answerable questions that include words frequently found in unanswerable queries. Additionally, conducting an N-gram analysis before deploying the data is crucial to identify patterns within unanswerable questions. We hope that our paper will inspire future benchmarks to consider these kinds of factors, thereby creating datasets free from data bias.

\section*{Social Impacts}

Our research addresses the critical issue of data bias in EHR QA benchmarks, particularly in the EHRSQL dataset, and offers a straightforward solution to mitigate this bias. The implications of this work are significant in enhancing the reliability and ethical use of AI in healthcare. By reducing data bias, our method contributes to the development of AI tools that support more accurate and equitable medical decision-making. This is crucial in healthcare, where data-driven insights directly impact patient care and outcomes. Furthermore, our work highlights the importance of ethical AI development and encourages further research in this area. Although our approach has limitations, specifically its applicability to EHRSQL, it sets a precedent for addressing similar challenges in other datasets, ultimately leading to more robust and unbiased AI applications in healthcare.

\bibliography{iclr2024_conference}
\bibliographystyle{iclr2024_conference}

\clearpage
\appendix

\section{Related Work}

\textbf{Semantic parsing} translates natural language into logical forms, including SQL queries. 
Several datasets exist to evaluate a model's ability to translate natural language into SQL.
Datasets such as WikiSQL \,\citep{zhong2017seq2sql} span diverse database domains, while Spider\,\citep{yu2018spider} and SParC\,\citep{yu2019sparc} introduce multi-table operations and multi-turn queries, respectively. KaggleDBQA\,\citep{lee2021kaggledbqa} aims for real-world authenticity with Kaggle databases, and SEDE\,\citep{hazoom2021text} addresses SQL queries from the Stack Exchange community.


\textbf{EHR QA} has seen significantly developed to automate the process of extracting essential information from EHRs.
Initially, emrQA~\citep{pampari2018emrqa} addresses clinical note question-answering, focusing on physicians' frequently asked questions. This evolved into emrKBQA~\citep{raghavan2021emrkbqa}, tailored to structured patient data in MIMIC-III. MIMICSQL~\citep{wang2020text} introduces innovative EHR Question Answering\,(QA) using text-to-SQL methods. This dataset employs rule-based automated question generation, later refined through crowd-sourcing. EHRSQL~\citep{lee2022ehrsql} enhances these datasets by incorporating practical questions and introducing unanswerable questions unseen during training, reflecting a more realistic real-world context.


\textbf{Unanswerable questions} are a recurring subject in natural language processing (NLP) \citep{rajpurkar2018know, kim-etal-2023-qa, min2022crepe, jia2017adversarial, sulem2022yes} and semantic parsing \citep{zhang2020did}, often included in training datasets. For instance, \citet{clark2017simple} employs paragraphs lacking direct answers, adjusting model confidence during training. \citet{devlin2018bert} introduces a specialized token for answer-less spans, while \citet{brown2020language} and subsequent studies \citep{zhu2021retrieving} allow models to generate text indicating the absence of an answer. Notably, most methods utilize unanswerable questions during training. 
In contrast, in our approach, unanswerable questions are intentionally included only in the validation and test sets, following the Out-of-Domain (OOD) approach of EHRSQL~\citep{lee2022ehrsql}.

\section{Uncertainty Estimation }
\label{appendix:uncertainty_estimation}

\paragraph{Beam Score} Beam decoding is a technique extensively used in various natural language processing (NLP) tasks, especially in areas of language generation such as semantic parsing. At each step of the decoding process, the algorithm selects the top-k candidate tokens that are most likely to produce subsequent tokens. This procedure is repeated for every step, calculating the probabilities associated with these k potential tokens. These probabilities are cumulatively added at each stage. The beam score is then derived by normalizing this cumulative sum, which can be expressed as follows:

$$score_{beam}(x, y_1, ..., y_t) = \frac{1}{t} \sum_{i=1}^{t}\log p(y_i|y_1, ..., y_{i-1}, x)$$

\noindent where $p$ represents the language model, $y_i$ denotes the generated token at each timestamp, and $x$ is the input. The sequence with the highest beam score is chosen, and this score is utilized to estimate uncertainty.

\paragraph{Maximum Entropy}

In language generation, predictive uncertainty quantifies the entropy of the token probability distributions that a model predicts\,\citep{xiao-wang-2021-hallucination}. Given the language model $p$, the entropy for each position is calculated as follows:

$$  H(y_i, c_i) =  - \sum_{v \in V}p(y_i = v |c_i) \log p(y_i = v |c_i) $$

\noindent where $c_i = \{x, y_1, ..., y_{i-1}\}$,  $V$ denotes the set of vocabularies, $y_i$ represents the generated token at each timestamp, and $x$ is the input. For a sequence of length $L$, there will be $L$ entropy values ranging from $H(y_1, c_1)$ to $H(y_L, c_L)$. Among these values, we select the maximum value of $H(y_i, c_i)$, thus calculating the uncertainty based on the most uncertain token.

\section{Implementation Details}
\label{appendix:implementation_details}


In this section, we delve into the implementation details of training the text-to-SQL model, generating SQL, and abstaining from unanswerable questions.

For the text-to-SQL model training, we set the learning rate to 1e-4 with no warmup steps or learning rate scheduler. We use a batch size of 4, with accumulation steps set at 8. The Adam optimizer is employed, and the gradient norm is clipped to a value of 1.0. Weight decay is configured at 0.1. All models are trained using a single NVIDIA RTX 3090. For experiments using the new split, we retrain the model because the validation set has changed.

In the evaluation step, where SQL is generated from questions, we employ beam decoding with a beam size of 5. The repetition penalty is set to 1.0, and early stopping of beam decoding is permitted. The maximum length is capped at 512. During generation, both beam score and entropy are computed.

For the process of filtering words based on N-gram-patterns and an uncertainty measure, we assess the ratio of N-gram patterns using the validation set. Patterns are extracted at three thresholds: for unigrams, bigrams, and trigrams. Initially, we filter out questions containing N-grams that have a ratio exceeding the set thresholds. Following this, we proceed to further refine the selection based on uncertainty estimates like the absolute beam score or the maximum entropy value. Predictions are set to SQL for questions deemed answerable and to a null value for those deemed unanswerable. Lastly, the generated SQL is executed on the MIMIC-III database to evaluate the performance of the actual execution.

\section{More Results}
\label{appendix:more_results}

In this section, we present additional experimental results that are not shown in Section \ref{sec:experiments}. Since, the utilization of test leaderboard is limited, we have conducted additional experiments using our new split constructed from the validation set of the original dataset. 

\subsection{Results on Random Split}
\label{appendix:random_split}

One might think that the results obtained using our proposed method are due to the reduced number of samples in the validation and test sets. To verify this, we conducted additional experiments using randomly split validation and test sets, both of equal size. The results are demonstrated in Table~\ref{tab:random_split}.

Clearly, we can observe that incorporating N-grams improves performance by nearly 20 points in the $F1_{exe}$ score for beam, and by nearly 30 for entropy. Therefore, we can conclude that the effectiveness of our proposed solution does not stem from the reduction in the size of the validation set, but rather from the removal of certain patterns in the validation set.

\begin{table}[t]
    \small
    \centering
    \caption{Results on the randomly split EHRSQL.}
    \resizebox{.7\columnwidth}{!}{
    
    \begin{tabular}{lcccc|cccc}
    \toprule
    \multirow{2}{*}{\textbf{Method}} & \multicolumn{4}{c|}{Valid} & \multicolumn{4}{c}{Test} \\
    \cmidrule{2-9}
    & \textbf{$F1_{ans}$} & \textbf{$P_{exe}$} & \textbf{$R_{exe}$} & \textbf{$F1_{exe}$} & \textbf{$F1_{ans}$} & \textbf{$P_{exe}$} & \textbf{$R_{exe}$} & \textbf{$F1_{exe}$} \\
    \midrule
    $Entropy$ & 31.0 & 100.0 & 18.4 & 31.0 & 35.6 & 94.3 & 21.6 & 35.2 \\
    $Beam$ & 42.0 & 100.0 & 26.6 & 42.0 & 44.3 & 97.4 & 28.7 & 44.3 \\
    $Ngram$ & 6.7 & 100.0 & 3.5 & 6.7 & 7.5 & 100.0 & 3.9 & 7.5 \\
    $Entropy + Ngram$ & 67.5 & 99.5 & 51.1 & 67.5 & 64.4 & 98.4 & 47.1 & 63.7 \\
    $Beam + Ngram$ & 63.8 & 99.4 & 46.5 & 63.4 & 63.2 & 98.9 & 46.1 & 62.9 \\
    \bottomrule
    \end{tabular}
    }
        \label{tab:random_split}
\end{table}

\subsection{Ablation on Different Data Splits}
\label{appendix:ablation_split}

We conduct an ablation study on the ratio threshold for data splitting. We tune three hyperparameters with different values: $\lambda$ for unigrams, $\lambda$ for bigrams, both of which are used for splitting questions, and $M$, which ensures the validation set includes specific patterns from the test set. We set these thresholds to balance the sizes of the validation and test sets, resulting in a higher $\lambda$ corresponding to a lower $M$. We experiments with three additional splits, ranging from 22-18-5 to 16-10-12. Beyond these values, increasing $\lambda$ and $M$ becomes meaningless, as it cannot maintain a balanced distribution between the validation and test sets.

The results are presented in Table \ref{tab:ablation_split}. For the entropy results in the 16-10-12 configuration, we use the value for the highest $P_{exe}$ because it is not feasible to meet the $99.0$ constraint for $P_{exe}$ using entropy value alone. As observed, the 22-18-5 configuration, which has a low $M$, indicating that the validation set is less susceptible to data bias, displays a similar trend where adding N-gram patterns does not notably enhance the $F1_{exe}$ performance for both beam and entropy. However, when $M$ is increased to 9, there is a marked change: introducing N-grams significantly boosts performance for both entropy and beam. Moreover, using N-gram alone yields better results compared to a low $M$ value of 5. This trend becomes even more pronounced in the 16-10-12 configuration, where the data bias remains unresolved. Therefore, it is required to set $M$ lower than 5 to resolve data bias for new split of validation set and test set.

\begin{table}[t]
\small
\centering
\caption{Ablation study on the hyperparameters used to construct the new split of the validation and test sets. The format X-Y-Z denotes X as the value of $\lambda$ for unigrams, Y as the value of $\lambda$ for bigrams, and Z as the value for $M$.}
\resizebox{.8\columnwidth}{!}{
    \begin{tabular}{lcccc|cccc} 
        \toprule
        \multirow{2}{*}{\textbf{Method}} & \multicolumn{4}{c|}{Valid} & \multicolumn{4}{c}{Test} \\
        \cmidrule{2-9} 
         & \textbf{$F1_{ans}$} & \textbf{$P_{exe}$} & \textbf{$R_{exe}$} & \textbf{$F1_{exe}$} & \textbf{$F1_{ans}$} & \textbf{$P_{exe}$} & \textbf{$R_{exe}$} & \textbf{$F1_{exe}$} \\
         \midrule   
         $\circ$ \ {\texttt{\,22-18-5}} & & & & \\
         $Entropy$ & 45.1 & 99.1 & 29.2 & 45.1 & 45.9 & 96.6 & 29.7 & 45.5 \\
         $Beam$ & 49.7 & 99.2 & 33.2 & 49.7 & 51.5 & 97.1 & 35.0 & 51.5 \\
         $Ngram$ & 0.5 & 100.0 & 0.3 & 0.5 & 1.6 & 100.0 & 0.8 & 1.6 \\
         $Entropy + Ngram$ & 53.0 & 99.3 & 35.8 & 52.6 & 51.7 & 94.2 & 34.2 & 50.2 \\
        $Beam + Ngram$ & 55.5 & 99.3 & 38.2 & 55.1 & 53.7 & 96.6 & 36.8 & 53.3 \\
        
        \midrule
        $\circ$ \ {\texttt{\,18-15-9}} & & & & \\
        $Entropy$ & 14.2 & 100.0 & 7.6 & 14.2 & 14.6 & 100.0 & 7.9 & 14.6 \\
         $Beam$ & 14.6 & 100.0 & 7.9 & 14.6 & 16.0 & 100.0 & 8.7 & 16.0 \\
         $Ngram$ & 10.0 & 100.0 & 5.3 & 10.0 & 10.9 & 90.9 & 5.3 & 10.0 \\
         $Entropy + Ngram$ & 47.4 & 99.2 & 30.8 & 47.0 & 42.7 & 99.0 & 26.8 & 42.2 \\
        $Beam + Ngram$ & 88.3 & 99.3 & 79.5 & 88.3 & 87.1 & 99.0 & 76.8 & 86.5 \\

        \midrule
        $\circ$ \ {\texttt{\,16-10-12}} & & & & \\
        $Entropy$ & 13.2 & 96.4 & 7.1 & 13.2 & 16.9 & 100.0 & 9.2 & 16.9 \\
         $Beam$ & 1.6 & 100.0 & 0.8 & 1.6 & 6.6 & 100.0 & 3.4 & 6.6 \\
         $Ngram$ & 10.5 & 100.0 & 5.5 & 10.5 & 7.1 & 92.9 & 3.4 & 6.6 \\
         $Entropy + Ngram$ & 69.3 & 99.5 & 53.2 & 69.3 & 64.0 & 99.4 & 46.8 & 63.7 \\
        $Beam + Ngram$ & 78.2 & 99.2 & 64.5 & 78.2 & 79.2 & 99.2 & 65.0 & 78.5 \\

         \bottomrule
    \end{tabular}
}
    \label{tab:ablation_split}
    \vspace{-5pt}
\end{table}

\subsection{Ablation on Filtering }
\label{appendix:filtering}

Table \ref{tab:filter_beam} and Table \ref{tab:filter_entropy} present an ablation study on the ratio threshold for N-gram filtering. We apply same threshold for filtering unigrams, bigrams, and trigrams. Setting the threshold lower implies filtering more questions using an increased number of N-gram patterns, whereas a higher threshold leans more towards uncertainty filtering. Notably, a very low threshold does not always result in better performance. This can be attributed to the inadvertent filtering of answerable questions that should be addressed. Typically, thresholds between 4 and 6 yield the best performance. We selects a threshold of 6 for our main experiments, as it produced the most balanced scores between entropy-based and beam-based methods.

Furthermore, thresholds above 6 yield results similar to those of 6. This occurs because there are no additional patterns beyond this point. Some N-grams with higher ratios have already been filtered out via uncertainty estimation, leaving the performance unchanged. Consequently, experimenting with thresholds higher than 7 is redundant, as the outcomes remain consistent.

\begin{table}[t]
\small
\centering
\caption{Ablation study on the threshold for N-gram filtering, conducted on the newly split set in conjunction with the beam score.}
\resizebox{.7\columnwidth}{!}{
    \begin{tabular}{lcccc|cccc} 
        \toprule
        \multirow{2}{*}{\textbf{Threshold}} & \multicolumn{4}{c|}{Valid} & \multicolumn{4}{c}{Test} \\
        \cmidrule{2-9} 
         & \textbf{$F1_{ans}$} & \textbf{$P_{exe}$} & \textbf{$R_{exe}$} & \textbf{$F1_{exe}$} & \textbf{$F1_{ans}$} & \textbf{$P_{exe}$} & \textbf{$R_{exe}$} & \textbf{$F1_{exe}$} \\
         \midrule  
         1 & 56.6 & 100.0 & 39.5 & 56.6 & 57.9 & 99.4 & 40.5 & 57.6 \\
         2 & 59.8 & 99.4 & 42.4 & 59.4 & 59.4 & 98.8 & 42.1 & 59.0 \\
         3 & 61.8 & 99.4 & 44.5 & 61.4 & 62.4 & 98.9 & 45.3 & 62.1 \\
        4 & 60.9 & 99.4 & 43.9 & 60.9 & 62.8 & 98.3 & 45.8 & 62.5 \\
        5 & 60.4 & 99.4 & 43.4 & 60.4 & 60.7 & 97.7 & 43.7 & 60.4 \\
        6 & 62.4 & 99.4 & 45.5 & 62.4 & 62.5 & 97.7 & 45.5 & 62.1 \\
        7 & 62.4 & 99.4 & 45.5 & 62.4 & 62.5 & 97.7 & 45.5 & 62.1 \\
        $\infty$ (baseline) & 59.4 & 99.4 & 42.4 & 59.4 & 60.3 & 98.2 & 43.2 & 60.0 \\
         \bottomrule
    \end{tabular}
}
    \label{tab:filter_beam}
\end{table}

\begin{table}[t]
\vspace{-5pt}
\small
\centering
\caption{Ablation study on the threshold for N-gram filtering, conducted on the newly split set in conjunction with the maximum entropy.}
\resizebox{.7\columnwidth}{!}{
    \begin{tabular}{lcccc|cccc} 
        \toprule
        \multirow{2}{*}{\textbf{Threshold}} & \multicolumn{4}{c|}{Valid} & \multicolumn{4}{c}{Test} \\
        \cmidrule{2-9} 
         & \textbf{$F1_{ans}$} & \textbf{$P_{exe}$} & \textbf{$R_{exe}$} & \textbf{$F1_{exe}$} & \textbf{$F1_{ans}$} & \textbf{$P_{exe}$} & \textbf{$R_{exe}$} & \textbf{$F1_{exe}$} \\
         \midrule 
         1 & 43.3 & 99.1 & 27.4 & 42.9 & 48.3 & 98.4 & 31.3 & 47.5 \\
         2 &59.0 & 99.4 & 41.6 & 58.6 & 58.9 & 98.1 & 41.3 & 58.2 \\
         3 & 53.5 & 99.3 & 36.6 & 53.5 & 57.1 & 100.0 & 40.0 & 57.1 \\
        4 & 53.7 & 99.3 & 36.8 & 53.7 & 57.9 & 99.4 & 40.5 & 57.6 \\
        5 & 57.3 & 99.4 & 40.3 & 57.3 & 59.6 & 98.2 & 42.4 & 59.2 \\
        6 & 57.3 & 99.4 & 40.3 & 57.3 & 59.6 & 98.2 & 42.4 & 59.2 \\
        7 & 57.3 & 99.4 & 40.3 & 57.3 & 59.6 & 98.2 & 42.4 & 59.2 \\
        $\infty$ (baseline) & 57.6 & 99.4 & 40.5 & 57.6 & 59.8 & 98.2 & 42.6 & 59.4 \\
         \bottomrule
    \end{tabular}
}
    \label{tab:filter_entropy}
\end{table}

\subsection{Ablation on Model Size}
\label{appendix:model_size}

Table \ref{tab:size_small} and Table \ref{tab:size_large} present the results of an ablation study on the model size. We use the same configuration for experiments on the new split as shown in Table \ref{tab:split}. As observed, with the new split, the model maintains a similar trend: adding N-gram filtering does not lead to a significant improvement in performance, especially in terms of entropy. It is also noteworthy that for T5-small, $P_{exe}$ decreases significantly on the test set when compared to T5-base and T5-large. This suggests that T5-small may lack generalization ability due to its smaller parameter count.


\begin{table}[t!]
\small
\centering
\caption{Performance results using the T5-small model on the new split.}
\vspace{5pt}
\resizebox{.75\columnwidth}{!}{
    \begin{tabular}{lcccc|cccc} 
        \toprule
        \multirow{2}{*}{\textbf{Method}} & \multicolumn{4}{c|}{Valid} & \multicolumn{4}{c}{Test} \\
        \cmidrule{2-9} 
         & \textbf{$F1_{ans}$} & \textbf{$P_{exe}$} & \textbf{$R_{exe}$} & \textbf{$F1_{exe}$} & \textbf{$F1_{ans}$} & \textbf{$P_{exe}$} & \textbf{$R_{exe}$} & \textbf{$F1_{exe}$} \\
         \midrule   $Entropy$ & 17.3 & 100.0 & 9.5 & 17.3 & 22.4 & 97.9 & 12.4 & 22.0 \\
         $Beam$ & 18.2  & 100.0 & 10.0 & 18.2 & 27.7 & 98.4 & 15.8 & 27.2 \\
         $Ngram$ & 2.1 & 100.0 & 1.1 & 2.1 &  1.6 & 100.0 & 0.8 & 1.6 \\
         $Entropy + Ngram$ & 17.3 & 100.0 & 9.5 & 17.3 & 21.6 & 97.8 & 11.8 & 21.1 \\
        $Beam + Ngram$ & 24.5 & 100.0 & 14.0 & 24.5 & 30.7 & 98.6 & 17.9 & 30.3 \\
         \bottomrule
    \end{tabular}
}
    \label{tab:size_small}
\end{table}

\begin{table}[t!]
\small
\centering
\caption{Performance results using the T5-large model on the new split.}
\vspace{5pt}
\resizebox{.75\columnwidth}{!}{
    \begin{tabular}{lcccc|cccc} 
        \toprule
        \multirow{2}{*}{\textbf{Method}} & \multicolumn{4}{c|}{Valid} & \multicolumn{4}{c}{Test} \\
        \cmidrule{2-9} 
         & \textbf{$F1_{ans}$} & \textbf{$P_{exe}$} & \textbf{$R_{exe}$} & \textbf{$F1_{exe}$} & \textbf{$F1_{ans}$} & \textbf{$P_{exe}$} & \textbf{$R_{exe}$} & \textbf{$F1_{exe}$} \\
         \midrule   $Entropy$ & 7.1 & 100.0 & 3.7 & 7.1 & 7.6 & 93.8 & 4.0 & 7.6 \\
         $Beam$ & 14.6 & 100.0 & 7.9 & 14.6 & 15.0 & 93.9 & 8.2 & 15.0 \\
         $Ngram$ & 2.1 & 100.0 & 1.1 & 2.1 &  1.6 & 100.0 & 0.8 & 1.6 \\
         $Entropy + Ngram$ & 7.1 & 100.0 & 3.7 & 7.1 & 7.6 & 93.8 & 4.0 & 7.6 \\
        $Beam + Ngram$ & 14.6 & 100.0 & 7.9 & 14.6 & 16.4 & 94.4 & 9.0 & 16.4  \\
         \bottomrule
    \end{tabular}
}
    \label{tab:size_large}
\end{table}

\end{document}